\documentclass[conference]{IEEEtran}

\usepackage[letterpaper,top=0.75in,bottom=1in,left=0.625in,right=0.625in]{geometry}
\usepackage{amsmath,amssymb,amsfonts}
\usepackage{algorithmic}
\usepackage{algorithm}
\usepackage{graphicx}
\usepackage{textcomp}
\usepackage{booktabs}
\usepackage{multirow}
\usepackage{xcolor}
\usepackage{cite}
\usepackage{url}
\usepackage{tikz}
\usetikzlibrary{positioning,arrows.meta,fit,calc}

\newcommand{\rev}[1]{#1}

\begin{document}

\title{FL-MAESTRO: Multi-Agent LLM Orchestration for Resource-Constrained Federated Learning}

\author{
\IEEEauthorblockN{Jiajun Wu\IEEEauthorrefmark{1}, Zirui Wang\IEEEauthorrefmark{1}, Jiayu Zhou\IEEEauthorrefmark{2}, Qiang Ye\IEEEauthorrefmark{1}, Steve Drew\IEEEauthorrefmark{1}}
\IEEEauthorblockA{
\IEEEauthorrefmark{1}Department of Electrical and Software Engineering, University of Calgary, Calgary, Canada\\
\IEEEauthorrefmark{2}School of Information, University of Michigan, Ann Arbor, MI, USA\\
Email: \{jiajun.wu1, zirui.wang1, qiang.ye, steve.drew\}@ucalgary.ca; jiayuz@umich.edu}
}

\maketitle

\begin{abstract}
In Federated Learning (FL), the communication topology is a runtime variable rather than a fixed design choice, since links and edge devices drop in and out during training. Each round, the server must commit three coupled decisions, namely the communication topology, per-client resource allocation, and the aggregation rule for combining local updates. Recent agentic systems have begun bringing large language models (LLM) into FL, but the existing line of work either operates at setup time or handles a single runtime dimension such as client selection. We propose \textit{FL-MAESTRO}, a multi-agent orchestrator that makes the joint runtime FL decision directly through three specialist LLM agents, one per decision dimension. A coordinator combines their analyses into a single decision, and a non-LLM feasibility check confirms it before the round executes. Because the orchestrator consumes the server's predicted-failure list, it withholds clients whose updates would never be aggregated, which removes the dominant source of wasted round energy in classical FL on volatile edge networks. Because client state is read as natural-text profiles, the same orchestrator extends to heterogeneous device classes without per-class energy models. On a non-IID CIFAR-10 benchmark, FL-MAESTRO matches the accuracy of the strongest energy-aware baseline while cutting wasted round energy from over a third to near zero. Code is available at \url{https://github.com/denoslab/FL-MAESTRO}.
\end{abstract}

\begin{IEEEkeywords}
Federated learning, large language models, internet of things, edge computing, network orchestration
\end{IEEEkeywords}

\section{Introduction}
\label{sec:intro}
\suppressfloats[t]

\begin{figure}[!t]
    \centering
    \includegraphics[width=\columnwidth]{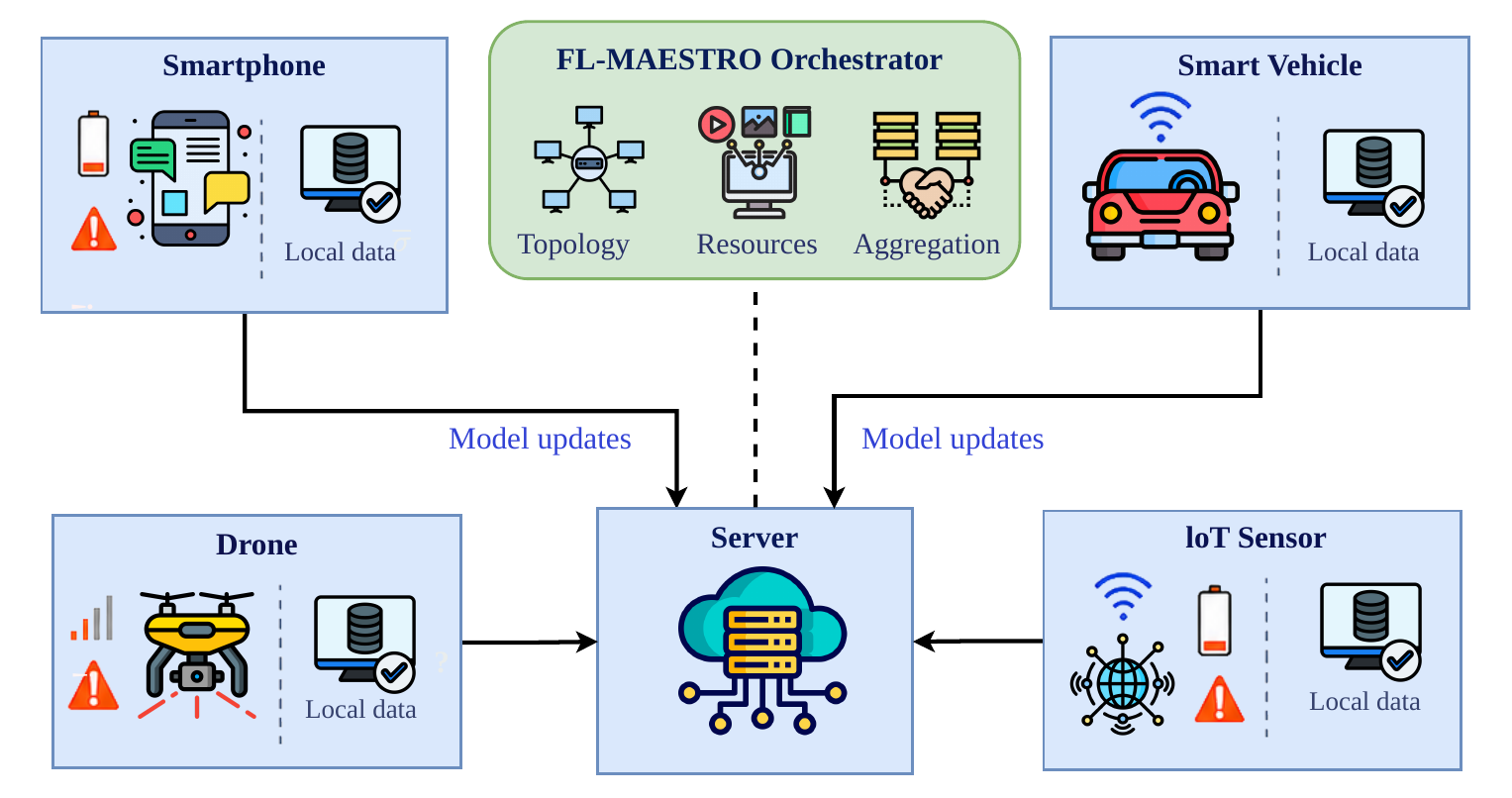}
    \caption{FL-MAESTRO deployment overview. The orchestrator coordinates topology, resource allocation, and aggregation decisions across clients.}
    \label{fig:deployment}
\end{figure}

Federated Learning (FL) trains a shared model across heterogeneous edge and IoT clients that operate under diverse power, computation, and communication budgets~\cite{lai2021oort, fedle2023}. Figure~\ref{fig:deployment} shows a typical FL deployment, where resource-constrained smartphones, smart vehicles, drones, and IoT sensors generate big data on-device and contribute to a central server. In each round, a FL central server selects available clients, broadcasts the current global model, lets each client perform local training on its private data, and aggregates the resulting local updates into a new global model.

Real edge deployments are dynamic. Clients drop out as their energy depletes or their connectivity falters, and client state varies widely across devices. Ideally the server would commit to a coupled set of decisions every round, namely the communication topology, per-client resource allocation, and aggregation weighting. Topology is the dimension most exposed to this volatility. Wireless links degrade under interference, mobility, and channel contention, so the connectivity graph a round runs on is not the one the server assumed when training began. A static star is unrealistic under these conditions, and hierarchical clustering, ring, or peer-to-peer alternatives may be the only feasible carriers when direct uplinks are saturated. The topology a round can use therefore depends on which clients survive it, which in turn depends on the resource allocation and aggregation rule chosen for that same round. Topology cannot be fixed offline or chosen in isolation, yet traditional FL algorithms do not consider these three decisions jointly. Making them together at runtime needs a decision-maker that can reason over heterogeneous client state and structured constraints.

Large language models (LLM) have grown beyond text generation into the reasoning core of agentic systems, which combine structured prompts with external tools to process mixed-modality state and emit structured decisions across many domains. FL can benefit from this kind of agentic decision-making, since its runtime control loop spans a configuration space larger than hand-coded rules can cover. Several recent works have begun exploring this direction. T-ELLM~\cite{tellm2025} uses an LLM for client selection while delegating resource allocation to a convex solver. Helmsman~\cite{helmsman2025} synthesizes FL system code at design time through collaborating LLM agents. Li et al.~\cite{agenticfl2025} sketch a broader vision of task-specialized agents automating the FL lifecycle, primarily at the strategy-design level, and leave the per-round runtime case open.

What makes the runtime case challenging is that the three decisions are tightly coupled. Allocating more local epochs to a battery-rich client improves accuracy only if the aggregation rule rewards that effort. Traditional FL methods such as FedAvg~\cite{mcmahan2017fedavg}, FedProx~\cite{li2020fedprox}, and FedNova~\cite{wang2020fednova} avoid this entanglement by fixing two dimensions, since they assume star aggregation, set every client to run same local epochs, and adapt only the aggregation weights. Resource-aware extensions like Oort~\cite{lai2021oort} adapt client selection but inherit the same fixed topology. As a result, no traditional FL method jointly considers topology, resources, and aggregation. The operational symptom is wasted round energy, where borderline-battery clients are admitted to a round, drain themselves on local training and uplink, then drop out before their update is aggregated. We call such clients predicted failures, since their depletion is computable from the client state. Three gaps motivate our work. First, no prior agentic system jointly orchestrates topology, resources, and aggregation while accounting for predicted client failures. Second, it is unsettled whether the heavy negotiation patterns borrowed from open-ended LLM benchmarks transfer to FL at all, since the runtime action space is small and feasibility is externally checkable. Third, no prior agentic FL system has been validated on open-weight, locally-hosted LLMs, leaving cloud dependency unaddressed for deployments where data must remain on-premises.

\begin{figure}[t]
    \centering
    \includegraphics[width=\columnwidth]{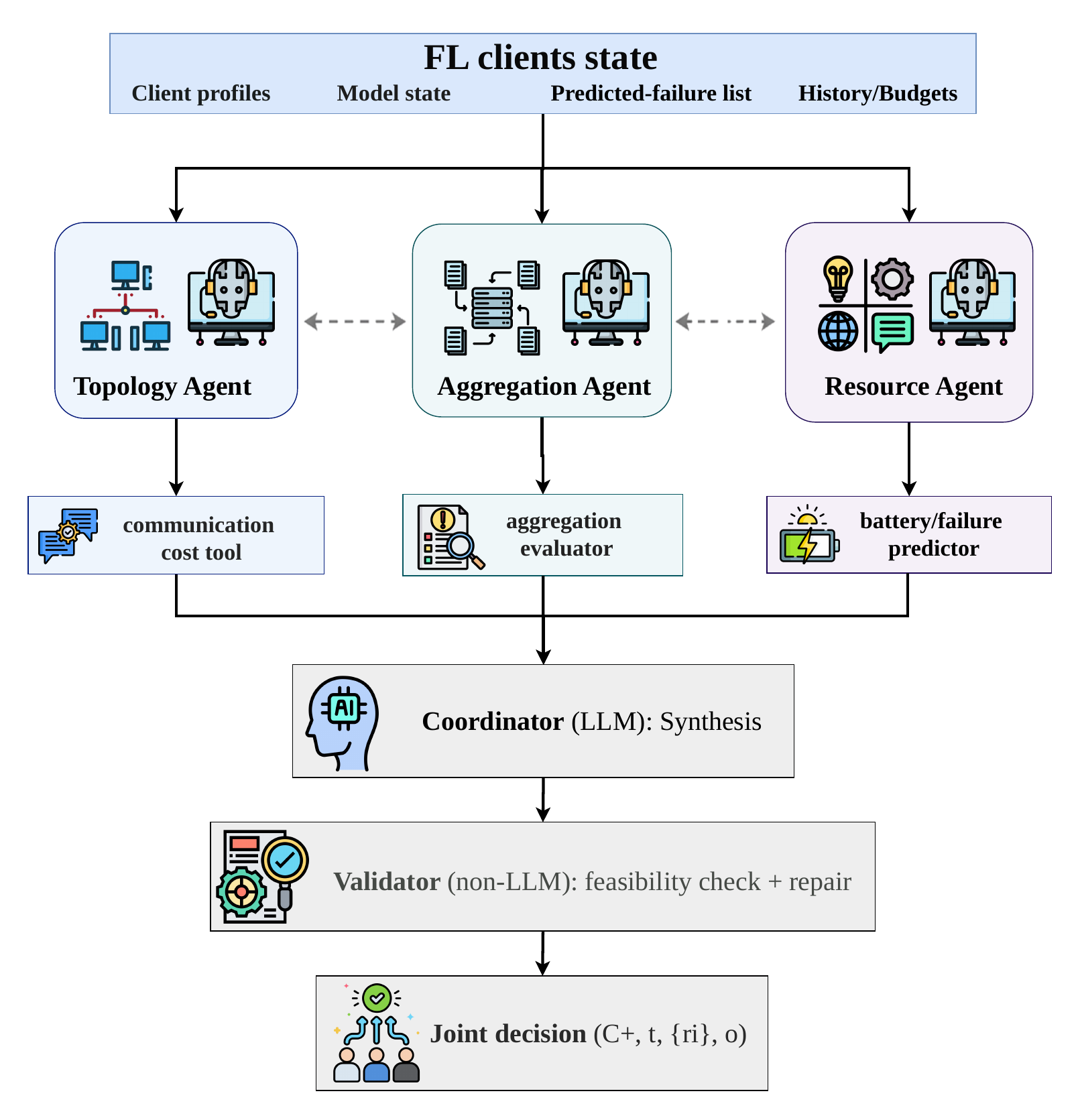}
    \caption{FL-MAESTRO architecture. Three specialist LLM agents read the FL state, query their domain tools, and exchange typed constraint analyses. The Coordinator synthesizes a joint decision and the Validator enforces feasibility.}
    \label{fig:architecture}
\end{figure}

To address these gaps, we propose FL-MAESTRO, a multi-agent orchestrator that makes the joint runtime FL decision every round through three specialist LLM agents, one per decision dimension. The TopologyAgent decides how clients communicate, the AggregationAgent decides how local updates are combined, and the ResourceAgent decides which clients participate and how much each one trains. Each agent grounds its reasoning in a domain tool, a coordinator synthesizes the three analyses into a single decision, and a non-LLM feasibility check verifies the result before the round executes. The orchestrator consumes the predicted-failure signal directly, holding borderline-battery clients out of the round instead of admitting them and burning their energy, which is the mechanism behind the energy savings reported below. Because client state is read as natural-text profiles, the same orchestrator also spans device classes that would otherwise need per-class energy models, and it runs on a self-hosted open-weight LLM, so cloud connectivity is not a hard requirement. Figure~\ref{fig:architecture} illustrates the architecture.

We summarize our technical contributions below:
\begin{itemize}
    \item We propose FL-MAESTRO, a multi-agent orchestrator that jointly orchestrates topology, resources, and aggregation each round through three specialist LLM agents and an external feasibility check, addressing the runtime joint decision case left open by FL work.
    \item On a non-IID FL benchmark, FL-MAESTRO drops wasted round energy from over a third to near zero while matching the accuracy of the strongest energy-aware baseline, with an ablation suggesting that lightweight agentic coordination suffices for this task.
    \item We show that even with a relatively small open-weight LLM such as Qwen3.5-35b, FL-MAESTRO matches the cloud-LLM results without a cloud-LLM dependency.
\end{itemize}

\section{Related Work}
\label{sec:related}

\subsection{FL Decision Subspaces}
Classical FL methods address the orchestration problem one decision dimension at a time. FedAvg~\cite{mcmahan2017fedavg}, FedProx~\cite{li2020fedprox}, FedNova~\cite{wang2020fednova}, and q-FedAvg~\cite{li2020qffl} fix the star topology and uniform local-epoch budget, varying only the aggregation weighting strategy. Resource-aware methods~\cite{wang2019adaptive, fedle2023} adapt per-client local epochs to runtime client state, and selection-aware methods like Oort~\cite{lai2021oort} bias selection toward high-utility clients, but both hold topology fixed. \rev{More recent work continues this single-axis pattern: Vardhan et al.~\cite{vardhan2025clientsel} optimize client selection jointly under data heterogeneity and network latency but leave topology and aggregation untouched, while Liu et al.~\cite{liu2025hatdfed} formulate heterogeneity-aware topology optimization for decentralized FL as a dual NP-hard problem yet decouple it from aggregation weighting and per-client workload.} Wu et al.~\cite{wu2024topologyfl} survey topology-aware FL and find that many designs still decouple topology from per-client resource allocation. None of these methods address the joint coupling across topology, per-client resources, and aggregation under failure-aware constraints.

\subsection{LLM Agents for FL Decisions}
Recent work applies LLM agents to FL setup and selection. T-ELLM~\cite{tellm2025} uses an LLM to generate device-selection policies for wireless FL while delegating resource allocation to convex optimization, addressing two dimensions through a hybrid pipeline. Helmsman~\cite{helmsman2025} synthesizes FL system code via multi-agent collaboration at design-time rather than per-round runtime. ComAgent~\cite{comagent2025} addresses wireless beamforming through a sequential multi-agent pipeline. Li et al.~\cite{agenticfl2025} sketch a vision of agents covering the FL lifecycle but leave open the operational question of how much agentic coordination is needed for runtime joint orchestration. FL-MAESTRO is the minimal-instantiation answer for that runtime case.

\section{System Model and Problem Formulation}
\label{sec:system}

\subsection{Federated Learning Setting}

Consider an FL system with clients $\mathcal{C} = \{c_1, \ldots, c_N\}$ training a global model $\mathbf{w}$ over $R$ communication rounds. Each client $c_i$ has a profile $\mathbf{p}_i = (b_i, \mu_i, \beta_i, d_i, \boldsymbol{\ell}_i, a_i)$: remaining battery $b_i \in [0,1]$, compute capability $\mu_i > 0$, bandwidth $\beta_i > 0$ (Mbps), local data size $d_i$, label distribution $\boldsymbol{\ell}_i$, and aggregation capability $a_i \in \{0,1\}$. Profiles are dynamic across rounds on three dimensions: battery depletion, the predicted-failure list $F^{(r)}$ recomputed each round from residual energy, and the consequent shrinkage of the eligible subset $\mathcal{C} \setminus F^{(r)}$. We model bandwidth as a per-client scalar $\beta_i$, and treat mobility, channel-quality fluctuation, mid-round handover, multi-user interference, OFDMA scheduling, and retransmission energy as constant or abstracted away within a round. Recent wireless-FL work~\cite{lai2021oort, fedle2023} layers explicit channel models on the same per-client allocation primitive we use, so those extensions compose with the orchestration layer. The unified-schema testbed isolates the orchestration mechanism, and Section~\ref{sec:hetero} discusses multi-class deployments.

\subsection{Three Coupled Decision Dimensions}

At each round $r$, the orchestrator commits to three coupled decisions over an active client subset $\mathcal{C}^+ \subseteq \mathcal{C} \setminus F^{(r)}$ where $F^{(r)}$ is the server's predicted-failure list at round $r$, meaning clients not in $F^{(r)}$ are battery-eligible and participate by default while flagged clients may be excluded by the orchestrator to prevent wasted energy on mid-round battery failures.
The predicted-failure list is computed deterministically from the FL state via $F^{(r)} = \{i : E_i(\tau_{\text{default}}, \mathbf{r}_{\text{default}}) > b_i\}$, where $\tau_{\text{default}} = \text{star}$ and $\mathbf{r}_{\text{default}} = (e_i = 5, \beta_i^{\text{alloc}} = \beta_i, \gamma_i = 1)$ are the default workload assumed by FedAvg-class baselines. $F^{(r)}$ therefore flags exactly the clients that would fail mid-round under unmodified FedAvg, making the comparison between $F$-aware and $F$-unaware orchestrators well-defined.

The communication topology is denoted by $\tau^{(r)} \in \mathcal{T} = \{\text{star}, \text{hier}, \text{ring}, \text{P2P}\}$ with per-round communication cost
\begin{equation}
\text{CommCost}(\tau, |\mathcal{C}^+|) =
\begin{cases}
2\,|\mathcal{C}^+|\,M\gamma & \tau = \text{star} \\
2\,|\mathcal{C}^+|\,M\gamma + 2AM\gamma & \tau = \text{hier} \\
2\,|\mathcal{C}^+|\,M\gamma & \tau = \text{ring} \\
k\,|\mathcal{C}^+|\,M\gamma & \tau = \text{P2P}
\end{cases}
\label{eq:commcost}
\end{equation}
where $M$ is the model byte-size, $\gamma \in (0,1]$ is the compression ratio, $A$ is the number of intermediate aggregators in the hierarchical case, and $k$ is the gossip neighborhood size in the P2P case. The factor of 2 counts uplink and downlink symmetrically. The hierarchical row assumes $A \ll |\mathcal{C}^+|$ with negligible intra-cluster overhead beyond the explicit aggregator-server term, and the P2P row counts one gossip round per FL round across $k$ neighbors per client. The per-client resource allocation is denoted by $\mathbf{r}_i^{(r)} = (e_i, \beta_i^{\text{alloc}}, \gamma_i)$ for each $i \in \mathcal{C}^+$, where $e_i$ is the number of local epochs, $\beta_i^{\text{alloc}}$ is the allocated bandwidth, and $\gamma_i$ is the per-client compression ratio. The aggregation weighting is denoted by $\sigma^{(r)} \in \{\text{uniform}, \text{data-size}, \text{loss-weighted}, \text{explicit}\}$ where the explicit case is parameterized by per-client weights $w_i \geq 0$ satisfying $\sum_{i \in \mathcal{C}^+} w_i = 1$.

\subsection{Coupling Constraints}

The three decisions are coupled through shared physical constraints. A client's per-round energy cost decomposes into a compute term and a communication term:
\begin{equation}
    E_i(\tau, \mathbf{r}_i) = \underbrace{\frac{e_i E_{\text{base}}}{\mu_i}}_{\text{compute}} + \underbrace{\frac{P_{\text{radio}} \cdot \text{CommCost}_i(\tau, \gamma_i)}{\beta_i^{\text{alloc}}}}_{\text{communication}}
    \label{eq:energy}
\end{equation}
where $E_{\text{base}}$ is the per-epoch baseline compute cost, $\mu_i$ scales compute cost inversely with the client's compute capability, $\text{CommCost}_i(\tau, \gamma_i)$ is client $i$'s share of $\text{CommCost}(\tau, |\mathcal{C}^+|)$ from Eq.~\ref{eq:commcost} adjusted for its per-client compression $\gamma_i$, $P_{\text{radio}}$ is a fixed radio draw coefficient, and $\beta_i^{\text{alloc}}$ is the bandwidth allocated to client $i$ in bytes/sec, so $\text{CommCost}_i / \beta_i^{\text{alloc}}$ gives transmission time, with values reported elsewhere in Mbps converted via $10^6/8$. Energy is expressed throughout as a fraction of a full battery, matching the normalized battery state $b_i \in [0,1]$, so the feasibility constraint $E_i \leq b_i$ compares like with like. That constraint implies whether client $i$ can participate at a given $\mathbf{r}_i$ depends on $\tau$, since $\tau$ determines $\text{CommCost}_i$ via Eq.~\ref{eq:commcost}. Hierarchical topologies add the constraint that aggregator clients in $\mathcal{C}^+$ must have $a_i = 1$, and ring topologies couple convergence to client ordering. Non-IID data further couples to the aggregation decision, since client updates are label-skewed. The orchestrator commits each round to a feasible joint decision $(\mathcal{C}^+, \tau, \{\mathbf{r}_i\}, \sigma)$ that respects these constraints, with a per-round decision space of cardinality $\mathcal{O}(2^{|\mathcal{C} \setminus F|} \cdot |\mathcal{T}| \cdot \prod_i |\mathcal{R}_i| \cdot 4)$ that is too large for exhaustive search.

\section{FL-MAESTRO Framework}
\label{sec:framework}

\subsection{Architecture Overview}

\rev{FL-MAESTRO uses three specialist LLM agents coordinated through the protocol in Fig.~\ref{fig:architecture}. Each agent receives the current FL state $\mathbf{s}^{(r)} = (\{\mathbf{p}_i\}, \mathbf{w}_{\text{state}}, B_{\text{state}}, H)$, containing client profiles, global model state, remaining budgets, and prior round decisions.}

\subsection{Specialist Agents}

\rev{The \textbf{TopologyAgent} proposes a topology type with configuration (e.g., aggregator assignments for hierarchical) and reports its communication cost. The \textbf{AggregationAgent} chooses among four schemes (uniform, data-size, loss-weighted, or explicit per-client weights) and, for hierarchical topology, the aggregator-to-child mapping; it consults a scheme-evaluation tool that materializes the candidate weight vectors before committing. The \textbf{ResourceAgent} proposes per-client local epochs, compression, and bandwidth, and verifies feasibility against energy-cost tools. All three agents share tools for communication cost, energy cost, label diversity, and battery feasibility, and the explicit-weight case generalizes to any normalized vector, so FedNova- and q-FFL-style weights are emitted without architectural change.}

\subsection{Coordinator and Validator}

\rev{The Coordinator uses the same LLM backend with a synthesis-only prompt and no tools, reading the three typed constraint-analysis records and emitting a feasible joint decision $(\mathcal{C}^+, \tau, \{\mathbf{r}_i\}, \sigma)$ as typed JSON. The Validator, a non-LLM Python function, enforces battery eligibility (Eq.~\ref{eq:energy}), aggregator capability ($a_i = 1$), weight normalization within $10^{-6}$, and $e_i \in [1, 20]$. Infeasible decisions are clipped to the feasibility boundary, so LLM failures surface as repairs rather than runtime crashes.}

\subsection{Collaboration Protocol}

\rev{Each FL round triggers a coordination cycle of up to $D_{\max} \in \{2, 3\}$ iterations. The three specialists emit initial proposals from the FL state, then revise them while seeing the other two dimensions, so coupling is enforced by exchange rather than by a single global proposer.}

\begin{algorithm}[t]
\caption{FL-MAESTRO Multi-Agent Collaboration}
\label{alg:collab}
\begin{algorithmic}[1]
\REQUIRE FL state $\mathbf{s}$ with predicted-failure list $F$, max analysis rounds $D_{\max}$
\STATE \textbf{Round 0} (initial analyses from FL state):
\STATE \quad TopologyAgent emits topology proposal $\hat{\tau}$
\STATE \quad AggregationAgent emits aggregation-scheme $\hat{\sigma}$
\STATE \quad ResourceAgent emits per-client allocation $\{\hat{\mathbf{r}}_i\}$
\FOR{$d = 1$ \textbf{to} $D_{\max} - 1$}
    \STATE TopologyAgent revises $\hat{\tau}$ given prior $\{\hat{\sigma}, \{\hat{\mathbf{r}}_i\}\}$
    \STATE AggregationAgent revises $\hat{\sigma}$ given prior $\{\hat{\tau}, \{\hat{\mathbf{r}}_i\}\}$
    \STATE ResourceAgent revises $\{\hat{\mathbf{r}}_i\}$ given prior $\{\hat{\tau}, \hat{\sigma}\}$
    \IF{all proposed topologies agree}
        \STATE \textbf{break}
    \ENDIF
\ENDFOR
\STATE Coordinator synthesizes $\mathcal{D} = (\mathcal{C}^+, \tau, \{\mathbf{r}_i\}, \sigma)$ from the three final constraint-analysis records
\STATE Validator enforces $\mathcal{C}^+ \subseteq (\mathcal{C}_{\text{eligible}} \setminus F)$, repair if violated
\RETURN Validated decision $\mathcal{D}$
\end{algorithmic}
\end{algorithm}

\section{Experimental Evaluation}
\label{sec:eval}

\subsection{Setup}

\textbf{Dataset and model.} CIFAR-10 with Dirichlet non-IID partitioning at $\alpha \in \{0.1, 0.3\}$, trained for up to 50 rounds with early stopping at 3 rounds patience and 0.001 minimum accuracy delta. The model is a CNN of 545,098 parameters, namely two $3\times3$ convolutional layers of 32 and 64 channels, each followed by ReLU and $2\times2$ max-pooling, then fully connected layers of 128 and 10 units. Clients train with SGD at learning rate 0.01, momentum 0.9, and batch size 32 under cross-entropy loss.

\textbf{Scenarios.} Two heterogeneous edge scenarios are evaluated, each with $N = 30$ clients. Per-client battery is sampled uniformly from $[0.5, 0.9]$, bandwidth is bimodal with 60\% of clients on WiFi at 5--20 Mbps and 40\% on cellular or IoT links at 1--5 Mbps, and compute capability is sampled from $\exp(\mathcal{N}(0.0,\, 0.5^2))$ clipped to $[0.3, 3.0]$. The \emph{hard} scenario uses Dirichlet $\alpha = 0.3$, so each client predominantly carries 2 to 3 of 10 classes, and forces 12 of 30 clients into a low-battery range $[0.12, 0.20]$. The \emph{extreme} scenario uses $\alpha = 0.1$, where each client carries predominantly 1 to 2 classes, and places 18 of 30 clients at low battery. Each strategy, scenario, and backend combination is evaluated on three random seeds controlling client profile sampling, data partitioning, and model initialization.

\textbf{Implementation parameters.} The per-client compression ratio defaults to $\gamma_i = 1.0$ unless modified by a strategy. The P2P gossip neighborhood size is fixed at $k = 3$. The hierarchical-topology aggregator count $A$ varies per round depending on which clients have $a_i = 1$ in their profile. Energy constants: $E_{\text{base}} = 0.005$ J, $P_{\text{radio}} = 0.001$ W, $M = 4$ MB.

\textbf{Baselines.} \emph{FedAvg}~\cite{mcmahan2017fedavg}: 5 epochs, star, data-size weighting. \emph{FedProx}~\cite{li2020fedprox}: $\mu = 0.01$. \emph{HierFixed}: hierarchical with first $a_i{=}1$ as aggregator and star fallback, 5 epochs, data-size. \emph{FedNova}~\cite{wang2020fednova}: per-client epochs by battery, $w_i \propto d_i / e_i$. \emph{q-FedAvg}~\cite{li2020qffl}: $w_i \propto \ell_i^2 \cdot d_i$, $q = 2$. \emph{FedLE}~\cite{fedle2023}: tolerance-aware epoch budgeting. \emph{ResourceAware}: epochs and compression scaled by battery and bandwidth. \emph{PipelineIndependent}: decoupled topology-then-resource pipeline. \emph{RuleBasedJoint}: hand-coded bandwidth/battery thresholds.

\textbf{Ablations.} \emph{FLM-Single}: single generalist LLM emits the full joint decision in one call. \emph{FLM-Independent}: three specialists, no cross-visibility, mechanically merged. \emph{FLM-Coupled} (primary system): three specialists with up to 3 analysis rounds, prior-round cross-visibility, coordinator synthesis, early exit on topology consensus. \emph{FLM-Coupled+A}: multi-threshold failure probing at three epoch-budget thresholds. \emph{FLM-Coupled+B}: conditional confidence-weighted negotiation round on detected conflicts. \emph{FLM-Coupled++}: A and B combined with up to 4 analysis rounds.

\textbf{Implementation.} All results come from simulation rather than a hardware testbed. FL training, client energy accounting, and battery depletion run in PyTorch on a single desktop workstation, with the CIFAR-10 CNN small enough that training does not require a GPU. Only LLM inference uses dedicated hardware.

\textbf{LLM backends.} Agents are implemented with CAMEL-AI~\cite{li2023camel}. Two backends are evaluated. The cloud backend is GPT-4.1-mini through API. The open-weight backend is Qwen3.5-35b served locally on an NVIDIA DGX Spark workstation with 128 GB unified memory, through Ollama.

\textbf{Metrics.} The primary metrics are best test accuracy across all training rounds, total wasted energy expressed as the fraction of round-aggregated energy spent on clients that failed mid-round, and per-round failure rate. We also report total communication cost in bytes and the average number of analysis rounds per FL round.

\subsection{Main Results}

\begin{figure}[t]
    \centering
    \includegraphics[width=0.8\columnwidth]{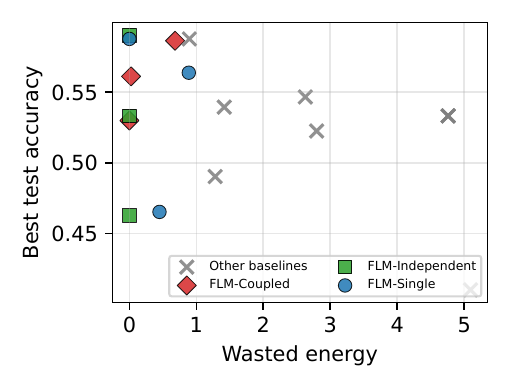}
    \caption{Energy-accuracy frontier on extreme. Each point is one (strategy, seed). FLM-Independent occupies the zero-waste corner. FLM-Coupled trades 0--4\% waste for the highest accuracy. FedAvg-class baselines cluster at 30\%+ waste.}
    \label{fig:pareto}
\end{figure}

FL-MAESTRO leads on both accuracy and energy. Table~\ref{tab:main} reports the head-to-head on the GPT-4.1-mini. FLM-Coupled tops accuracy on both scenarios while holding wasted round energy to a few percent. FLM-Independent is the zero-waste extreme, trading a small accuracy gap for exact zero waste and zero failures, a clean efficiency-favoring point on the design frontier. Classical baselines that do not consume the predicted-failure list cluster an order of magnitude higher on wasted energy because they admit borderline-battery clients, and the resource-aware heuristic closes part of that gap but trails the joint-orchestration variants. Fig.~\ref{fig:pareto} visualizes the (waste, accuracy) frontier. Orchestration runs as an asynchronous overlay between rounds, with a per-round window of $20$ to $90$ seconds across the two backends that fits within the typical inter-round gap at $N = 30$ where client upload and server-side aggregation dominate the wall-clock. Coordination converges in 2.0 analysis rounds per FL round on average across all seeds and scenarios, and orchestration runs entirely server-side, so it adds no client-side computation or communication.

\begin{table*}[t]
\centering
\caption{Main comparison on GPT-4.1-mini. Cells report best accuracy, wasted energy, and failure rate. USD/run measured from OpenAI usage logs on \texttt{gpt-4.1-mini-2025-04-14}. Best per column in bold.}
\label{tab:main}
\begin{tabular}{lccc|ccc|cc}
\toprule
& \multicolumn{3}{c|}{\textbf{Hard} ($\alpha = 0.3$)} & \multicolumn{3}{c|}{\textbf{Extreme} ($\alpha = 0.1$)} & & \\
\textbf{Method} & best\_acc & waste \% & fail \% & best\_acc & waste \% & fail \% & \textbf{calls/run} & \textbf{USD/run} \\
\midrule
FedAvg~\cite{mcmahan2017fedavg}     & 0.639 & 33.5 & 31.6 & 0.581 & 42.6 & 35.8 & --- & --- \\
FedProx~\cite{li2020fedprox}        & 0.640 & 33.5 & 31.6 & 0.580 & 42.6 & 35.8 & --- & --- \\
FedNova~\cite{wang2020fednova}      & 0.653 & \phantom{0}9.7 & 21.7 & 0.571 & 11.9 & 22.4 & --- & --- \\
q-FedAvg~\cite{li2020qffl}          & 0.613 & 26.9 & 23.0 & 0.518 & 29.8 & 21.9 & --- & --- \\
FedLE~\cite{fedle2023}       & 0.637 & 24.1 & 31.8 & 0.580 & 18.9 & 29.4 & --- & --- \\
HierFixed                   & 0.639 & 33.5 & 31.6 & 0.581 & 42.6 & 35.8 & --- & --- \\
ResourceAware                       & 0.651 & \phantom{0}7.0 & 11.4 & 0.598 & \phantom{0}6.9 & \phantom{0}9.2 & --- & --- \\
PipelineIndependent                 & 0.642 & 10.0 & 16.7 & 0.568 & 11.9 & 21.2 & --- & --- \\
RuleBasedJoint                      & 0.645 & 17.6 & 22.4 & 0.587 & 21.6 & 26.4 & --- & --- \\
\midrule
FL-MAESTRO-Single      & 0.655 & \phantom{0}2.3 & \phantom{0}1.9 & 0.606 & \phantom{0}2.1 & \phantom{0}1.4 & 18 & \$0.05 \\
FL-MAESTRO-Independent & 0.651 & \textbf{\phantom{0}0.0} & \textbf{\phantom{0}0.0} & 0.589 & \textbf{\phantom{0}0.0} & \textbf{\phantom{0}0.0} & 49 & \$0.13 \\
FL-MAESTRO-Coupled     & \textbf{0.657} & \phantom{0}3.9 & \phantom{0}3.5 & \textbf{0.615} & \phantom{0}0.2 & \phantom{0}0.2 & 109 & \$0.28 \\
\bottomrule
\end{tabular}
\end{table*}

\subsection{Robustness across seeds and backends}

The accuracy lead on extreme is sign-consistent across three seeds and reproduces on the open-weight Qwen3.5-35b backend, with Table~\ref{tab:multiseed} reporting three-seed means. A paired one-sided $t$-test gives $t = 0.32$ ($p = 0.39$) and a 95\% confidence interval just crossing zero, so we report this as directional consistency rather than statistical significance and leave an extended-seed study as future work. On both backends the qualitative ordering of FLM-Coupled / -Independent / -Single is preserved, and FLM-Independent's zero-waste and zero-failure invariants hold across all six three-seed cells. Qwen3.5 staying close to GPT-4.1-mini suggests smaller open-weight models can run FL-MAESTRO at cloud-LLM performance.

\begin{table}[t]
\centering
\caption{Three-seed mean for FL-MAESTRO variants. Each cell shows best\_acc, waste\%, fail\%.}
\label{tab:multiseed}
\begin{tabular}{lcc}
\toprule
\textbf{Variant} & \textbf{GPT-4.1-mini} & \textbf{Qwen3.5} \\
\midrule
\multicolumn{3}{l}{\textit{Hard} ($\alpha = 0.3$)} \\
FLM-Single      & $0.651$, $2.4$, $1.6$ & $0.646$, $0.6$, $5.5$ \\
FLM-Independent & $0.641$, $0.0$, $0.0$ & $0.641$, $0.0$, $0.0$ \\
FLM-Coupled     & $0.648$, $4.3$, $4.4$ & $0.651$, $0.0$, $0.2$ \\
\midrule
\multicolumn{3}{l}{\textit{Extreme} ($\alpha = 0.1$)} \\
FLM-Single      & $0.583$, $1.8$, $1.0$ & $0.573$, $0.7$, $3.9$ \\
FLM-Independent & $0.580$, $0.0$, $0.0$ & $0.552$, $0.0$, $0.0$ \\
FLM-Coupled     & $0.600$, $4.1$, $5.4$ & $0.572$, $0.6$, $2.0$ \\
\bottomrule
\end{tabular}
\end{table}

\subsection{Minimal coordination suffices for runtime FL decisions}

Adding negotiation or multi-threshold failure probing on top of FLM-Coupled yields no measurable accuracy gain on this testbed. The structural ablation in Table~\ref{tab:ablation} addresses Li et al.'s~\cite{agenticfl2025} open question on how much agentic coordination is needed. The task's structure explains why. The action space is typed JSON, an external validator checks feasibility, and tool outputs supply concrete cost estimates, leaving little uncertainty for added negotiation to resolve. Multi-agent debate~\cite{du2024debate} mainly helps in open-ended natural-language tasks without such verifiers. Multi-threshold failure probing (+A) directionally trades accuracy for safety, which we report as a Pareto shift rather than an established trade-off given $N = 3$. We are careful not to over-attribute on this controlled testbed, where simpler failure-aware baselines that consume $F^{(r)}$ would attain comparable round-energy efficiency.
\begin{table}[t]
\centering
\caption{Structural ablation. Three-seed mean $\pm$ SD on GPT-4.1-mini for FLM-Coupled-family variants. Best results in bold.}
\label{tab:ablation}
\begin{tabular}{lccc}
\toprule
\textbf{Variant} & \textbf{best\_acc} & \textbf{waste \%} & \textbf{fail \%} \\
\midrule
\multicolumn{4}{l}{\textit{Hard} ($\alpha = 0.3$)} \\
FLM-Coupled    & $\mathbf{0.648 \pm 0.011}$ & $4.3 \pm 1.4$ & $4.4 \pm 1.8$ \\
FLM-Coupled+A  & $0.645 \pm 0.009$           & $6.2 \pm 0.8$ & $5.8 \pm 3.3$ \\
FLM-Coupled+B  & $0.644 \pm 0.014$           & $5.0 \pm 1.2$ & $3.5 \pm 0.4$ \\
FLM-Coupled++  & $0.640 \pm 0.014$           & $3.0 \pm 0.4$ & $2.2 \pm 1.6$ \\
\midrule
\multicolumn{4}{l}{\textit{Extreme} ($\alpha = 0.1$)} \\
FLM-Coupled    & $\mathbf{0.600 \pm 0.022}$ & $4.1 \pm 2.8$ & $5.4 \pm 3.7$ \\
FLM-Coupled+A  & $0.585 \pm 0.021$           & $3.4 \pm 1.2$ & $1.9 \pm 1.1$ \\
FLM-Coupled+B  & $0.583 \pm 0.020$           & $3.4 \pm 2.7$ & $4.3 \pm 4.0$ \\
FLM-Coupled++  & $0.593 \pm 0.024$           & $3.9 \pm 1.4$ & $2.9 \pm 1.3$ \\
\bottomrule
\end{tabular}
\end{table}

\subsection{Heterogeneous-schema deployability}
\label{sec:hetero}

FL-MAESTRO's architecture also targets heterogeneous-schema deployments, where a filter over $F^{(r)}$ would require per-class customization. The controlled testbed isolates the orchestration mechanism by drawing all clients from a single profile schema, so this subsection is an architectural argument rather than a measured result. In real deployments, clients span device classes whose profile formats, energy models, and failure modes do not share a schema, and a hand-coded filter then requires per-class energy predictors, per-class failure-mode taxonomies, and a schema-unification layer. FL-MAESTRO avoids this engineering by reading profiles as natural-text records and composing the validator's feasibility predicate class-agnostically across device classes. Adding a new device class becomes a prompt update rather than a re-engineering of the orchestration logic. The cross-backend reproduction on Qwen3.5-35b retains this property under data-residency constraints, with self-hosted orchestration viable on workstation-class hardware. Scaling beyond the 30-client controlled testbed remains open, since the prompt-context budget likely limits direct application at thousands of clients without hierarchical decomposition.

\section{Conclusion}
\label{sec:conclusion}

We presented FL-MAESTRO, a multi-agent orchestrator that makes the joint runtime FL decision through three specialist LLM agents and an external feasibility check. By design, a new device class becomes a prompt update rather than per-class re-engineering, an architectural advantage we analyze but do not evaluate here. Cross-backend reproduction on the open-weight Qwen3.5-35b confirms operability under data-residency constraints. We leave extending FL-MAESTRO to larger deployments as future work.

\section*{Acknowledgment}
This work was supported by the NSERC Alliance - Alberta Innovates Advance Program ALLRP 602246-24 and the NSERC Discovery Grant RGPIN-2024-03954.

\bibliographystyle{IEEEtran}
\bibliography{references}

\end{document}